# Near-Optimal Hardware Design for Convolutional Neural Networks

Byungik Ahn
*Neurocoms Inc.*
Seoul, Korea
jerryahn@neurocoms.com

*Abstract*—Recently, the demand of low-power deep-learning hardware for industrial applications has been increasing. Most existing artificial intelligence (AI) chips have evolved to rely on new chip technologies rather than on radically new hardware architectures, to maintain their generality. This study proposes a novel, special-purpose, and high-efficiency hardware architecture for convolutional neural networks. The proposed architecture maximizes the utilization of multipliers by designing the computational circuit with the same structure as that of the computational flow of the model, rather than mapping computations to fixed hardware. In addition, a specially designed filter circuit simultaneously provides all the data of the receptive field, using only one memory read operation during each clock cycle; this allows the computation circuit to operate seamlessly without idle cycles. Our reference system based on the proposed architecture uses 97% of the peak-multiplication capability in actual computations required by the computation model throughout the computation period. In addition, overhead components are minimized so that the proportion of the resources constituting the non-multiplier components is smaller than that constituting the multiplier components, which are indispensable for the computational model. The efficiency of the proposed architecture is close to an ideally efficient system that cannot be improved further in terms of the performance-to-resource ratio. An implementation based on the proposed hardware architecture has been applied in commercial AI products.

*Keywords*— Artificial Intelligence, Deep Learning, Computer Architecture, Efficiency, Neuron Machine

## I. Introduction

In the era of the fourth industrial revolution, artificial intelligence (AI) technology has gained wide-spread use in industry. The efficiency of AI hardware is thus becoming an increasingly important issue. Efficient AI hardware can be used for a wide range of products ranging from CCTV cameras to self-driving cars in the form of distributed, small, low-power devices. The deep-learning chipset market is expected to reach $91.2 billion mark by 2025, due to the potential compounded annual growth rate of 45.2%, during the period between the years, 2019 and 2025 [1]. Consequently, global companies are competing to provide better AI chip solutions. These solutions rely on the advancements in semi-conductor technology and the availability of massive amounts of resources. However, semiconductor-integrated technology has started facing physical limitations as node sizes are approaching the size of silicon molecules [28]. Furthermore, alternative technologies such as quantum computing are in their infancy [2].

Improving the efficiency of the AI computational circuits could be a solution to these problems. Existing AI solutions have considerable potential for improvement in terms of architectural efficiency. For example, NVIDIA's Xavier, one of the most advanced AI chips, computes the SSD300/MobileNetV1 (SSD/MobileNet) [3] object detection convolutional neural network (CNN) model at a speed of 665 frames per second (fps) [4]. However, this result means that the chip utilizes only 5% of its peak capability of 16 trillion multiplications per second [5] because 665 inferences for the model actually require 818 billion multiplications. Moreover, the chip has approximately 250 times more transistors than the minimum required number to implement the same number of multipliers. As another example, Google Coral Edge uses hardware capable of 1.97 trillion multiplications per second with 4,096 multipliers and 480 MHz of clock speed [6]. However, because the chip can compute MobileNetV1 inference at 143 times per second [7], the effective multiplication speed is 81 billion per second, and the utilization rate is only 4.1%. These low efficiencies may be due to the fact that they are designed as general-purpose chips that can run any algorithm. By limiting the use of such chips to CNN computations, the systems can be simplified and the efficiencies can be substantially improved.

This work presents a novel computational architecture, or design method, specifically for CNN systems. The proposed architecture maximizes the utilization of multipliers by designing the circuit with the same structure as that of the computational flow of the computation model, rather than mapping computations to a fixed hardware. In addition, a distributed memory system and specially designed filter circuit provides all the receptor data to the computation circuits every clock cycle continuously throughout the entire computation period. A reference system based on the proposed architecture utilizes 97% of its peak multiplication capability, and 56% of total resources are used for multipliers , which are indispensable for the computational model. As a result, the proposed architecture achieves an efficiency close to optimal efficiency that cannot be improved further. The proposed architecture has been realized in commercial products.

The remainder of this paper is structured as follows. Section 2 describes the computation of CNN, and Section 3 presents the proposed CNN hardware architecture. Section 4 evaluates the results of the reference implementation. Finally, Sections 5, 6, and 7 present the commercialization examples, discussions, and the conclusions, respectively.

## II. CNN

CNN is the most widely used computational model in AI technology. It is used not only for processing visual information but also for natural language processing [24], speech recognition [25], and the AlphaGo Go player [8].

To recognize the input images, a CNN sequentially applies $k \times k$ convolution filters to generate a predetermined number of feature maps for each layer [9]. The feature maps of the $L^{th}$ CNN layer are used to compute the feature maps of the $(L + 1)^{th}$ layer. The value at the $x$, $y$ position of the $f^{th}$ feature map in the $(L + 1)^{th}$ layer can be computed as follows:

$$M_{xy}^f(L+1) = f\left(\sum_{c=0}^{F_L} \sum_{j=-\lfloor k/2 \rfloor}^{\lfloor k/2 \rfloor} \sum_{i=-\lfloor k/2 \rfloor}^{\lfloor k/2 \rfloor} M_{(s \cdot x+i)(s \cdot y+j)}^c(L) \times w_{ij}^c(L)\right) \quad (1)$$

where $F_L$, $k$, and $s$ denote the number of feature maps in layer $L$, the size of the filter, and the stride size, respectively. The complexity of the CNN computational model is often expressed in terms of the amount of multiplications required, which is derived from this equation. The total number of multiplications in layer $L$ can be derived from Equation 1, and is given by

$$CountMul_L \cong F_L \times F_{L+1} \times W \times H \times k \times k \quad (2)$$

where $W$ and $H$ are the horizontal and vertical dimensions of the feature map.

## III. PROPOSED HARDWARE ARCHITECTURE

### A. Neuron Machine Architecture

The proposed CNN hardware architecture is based on the artificial neural network hardware architecture known as neuron machine [10]. The neuron machine architecture comprises a hardware neuron (HN) that performs neural computations, and a memory part (MP) that supplies data to the HN as shown in Figure 1-a. The output of the MP becomes the input of the HN and vice versa. The HN has the same circuit structure as the data flow of model neurons and is composed of fully pipelined circuits to produce results every clock cycle. The MP seamlessly provides input data to the HN every clock cycle, simultaneously storing the newly computed neuron values.

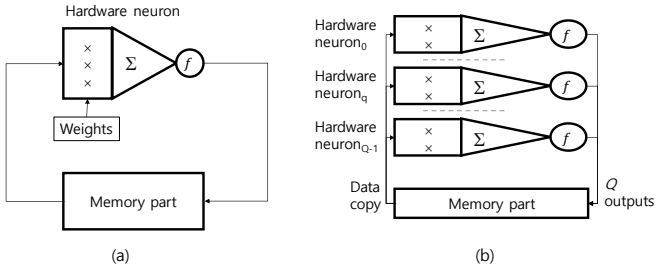

Fig. 1. (a) Neuron machine architecture. (b) System with multiple HNs that compute multiple feature maps in parallel.

### B. Conversion from Computational Model to Circuit

The CNN inference computation defined in Equation 1 can be expressed as a data-flow graph (DFG) [11]. The DFG that computes the value of a specific position $(x, y)$ of the convolution output feature map consists of $C \times k \times k$ multiplication nodes that multiply each of the values from the receptive fields at input position $(s \cdot x, s \cdot y)$ of all $C$ input feature maps by the corresponding weight values. A tree composed of addition nodes sums up all the multiplication results, and nodes for bias, a nonlinear function, and pooling are sequentially connected to the output of the addition tree.

The HN circuit is derived from the DFG, and the multiplication nodes in the graph are translated to pipelined multipliers with weight memories storing weight values, the addition tree is to a tree of pipelined adders, and nodes such as nonlinear functions are converted into circuits having corresponding functions. The difference between the DFG and the translated circuit is that the circuit uses $P \times k \times k$ multipliers instead of $C \times k \times k$, where $P$ is a fixed number and the multipliers and the adder tree process $P$ input feature maps simultaneously. In addition, an accumulator and a Netsum memory are placed at the end of the adder tree, so that the $\lceil C / P \rceil$ results of the feature maps are summed up sequentially, as shown in Figure 2.

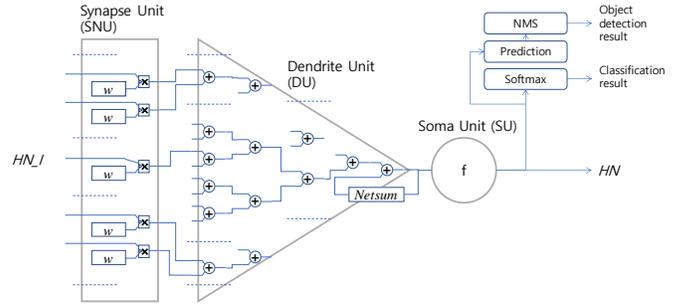

**Fig. 2. Block diagram of the HN illustrating it as a fully pipelined circuit with synapse unit (SNU), dendrite unit (DU), and soma unit (SU) connected in sequence.**

This simple circuit can seamlessly compute a large number of multiplications every clock cycle. In this way, by fitting the circuit to the computation model, it is possible to perfectly map the computational circuit with the computation. This is different from fixed hardware architectures such as the systolic array and vector processors, in which the mapping of circuits to computational models significantly affects the performance of the system.

### C. Input and Output of HN

In computing the $f^{th}$ feature map of a convolution layer, where the size of the convolution filter is $k \times k$ and the horizontal and vertical dimensions of the input feature map are $W$ and $H$, respectively, the input of the HN at the $t^{th}$ clock cycle after the computation begins can be described with a set of $k \times k \times P$ elements that can be defined as follows:

$$HN\_I(t) = \left\{ d_{xy}^c \middle| \begin{array}{l} c = \left\lfloor \frac{t}{W \times H} \right\rfloor \times P + (0 \ldots P - 1), \\ x = (t\%(W \times H))\%W + dw, \\ y = \left\lfloor \frac{t\%(W \times H)}{W} \right\rfloor + dy, \\ dx = -\left\lfloor \frac{k}{2} \right\rfloor \ldots \left\lfloor \frac{k}{2} \right\rfloor, \\ dy = -\left\lfloor \frac{k}{2} \right\rfloor \ldots \left\lfloor \frac{k}{2} \right\rfloor \end{array} \right\} \quad (3)$$

Where,

$$d_{xy}^c = \begin{cases} 0, & x < 0 \text{ or } y < 0 \text{ or } x \geq W \text{ or } y \geq H \\ M_{xy}^c(L), & \text{else} \end{cases} \quad (4)$$

and $M_{xy}^c(L)$ is the value of the $x, y$ position of the $c^{th}$ feature map of the layer $L$ described in Equation 1. In simpler terms, if $P = 1$, the input provided to the HN for each clock cycle constitutes all the data in the receptive field of the input image and is provided in the following order:

$$D_{0 \cdot 0}^0 \ldots D_{W-1 \cdot H-1}^0 \ldots D_{0 \cdot 0}^{C-1} \ldots D_{W-1 \cdot H-1}^{C-1}$$

Here, $D_{x \cdot y}^c$ is a data set consisting of $k \times k$ elements of the receptive field at the position $x, y$ of the $c^{th}$ input image. The same data sequence is repeated to compute all the $F$ output images.

The output of the HN is defined as follows.

$$HN(t') = M^f_{t'\%W' \cdot \lfloor t'/W' \rfloor}(L+1) \quad (5)$$

where $t'$ is the clock cycle sequence starting at a certain point in the middle of the computation and $W'$ is the width of the output feature map.

*D. Structure and Function of the HN*

First, each of the $k \times k \times P$ input values of the HN is multiplied with the corresponding weight (parameter) values in the synapse unit (SNU). These results serve as the outputs of the SNU. The weights are stored in $k \times k \times P$ distributed memories in the SNU. The output of the SNU is summed by the adder tree in the dendrite unit (DU). The adder tree adds $k \times k \times P$ values, and the accumulator at the end of the DU sequentially accumulates $\lceil C/P \rceil$ partial sums every $W \times H$ clock cycles, stores the $W \times H$ partial sums in memory, and outputs the accumulated result while processing the last feature maps. The output of the DU is supplied to the soma unit (SU) and is sequentially applied to the bias computation, activation function, and pooling. It finally results in the feature maps of the next layer, which is the output of the HN.

The output of the HN is sent to the MP and stored. In the final convolution layer, the computation result is applied to the softmax function, when the CNN model is used for classification, and to the prediction layer and non-maximum suppression (NMS) when it is used for object detection. The post-processing after the final layer is computed independently of the convolution computation. Thus, the first convolution layer for the next image can be started immediately after the last layer of the previous image.

Using pipeline registers, the entire circuit is pipelined such that one HN can process $k \times k \times P$ multiplications simultaneously every clock cycle throughout the whole computation period. However, this requires that input data to be supplied seamlessly every clock cycle.

*E. Memory Part*

The role of the MP is to continuously provide data to the HN in every clock cycle. The HN outputs the feature map data of layer $L$ and simultaneously stores the data of layer $L + 1$. The MP is composed of a memory array unit (MAU) for storing data and a receptor unit (RU) for converting data as shown in Figure 3.

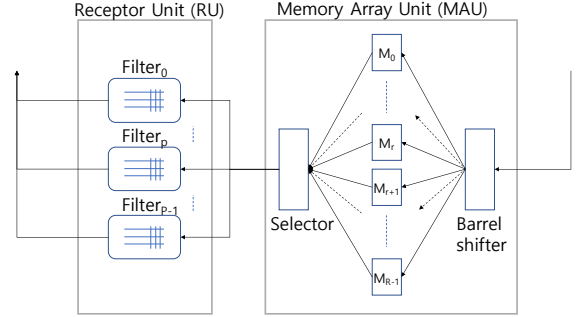

Fig. 3. Block diagram of MP. The MAU comprises of a large number of memories for storing inter-layer feature maps, and the receptor unit transforms the data.

**Memory Array Unit:** It is comprised of $R$ dual-port memories, where $R > P$. Dual-port memory is a type of memory that can simultaneously read and write in each clock cycle. A barrel shifter is placed between the input of the MAU and the write ports of the memories so that input data can be stored in memories arbitrarily according to the control signal. In addition, a selector is connected between the read ports of the memories and the output of the MAU, so that arbitrary $P$ consecutive memories can be selected for the data output. More specifically, the $f^{th}$ output feature map is stored in the $(f \% R)^{th}$ memory. In addition, at the $t^{th}$ clock cycle after the computation begins, data selected by the selector is can be represented by the following set with $P$ elements:

$$MAU(t) = \{M_{xy}^c | c = \lfloor t/(W \times H) \rfloor \times P + (0 \ldots P-1), x = t\%W, y = \lfloor t/W \rfloor\} \quad (6)$$

Here, $c$ identifies a group of $P$ input feature maps. It can also be described by the following sequence of data, each with $P$ elements.

$$\ldots \{M_{0 \cdot 0}^0, \ldots, M_{0 \cdot 0}^{P-1}\} \ldots \{M_{W-1 \cdot H-1}^0, \ldots, M_{W-1 \cdot H-1}^{P-1}\}$$
$$\{M_{0 \cdot 0}^P, \ldots, M_{0 \cdot 0}^{2P-1}\} \ldots \{M_{W-1 \cdot H-1}^P, \ldots, M_{W-1 \cdot H-1}^{2P-1}\} \ldots$$

**Receptor Unit:** As described in Equation 3, the quantity of data to be supplied as input of the HN every clock cycle is $k \times k \times P$, and if all of them are to be read from the memory, the memory bottleneck problem would occur. The RU converts $P$ data per clock cycle into $k \times k \times P$, to reduce the memory bandwidth. The RU consists of $P$ filter circuits called receptors, each of which converts one data flow per clock cycle to $k \times k$.

The receptor consists of $k$ shift register arrays each with $W$ registers connected in sequence, as shown in Figure 4-a. The values of the input feature maps are input sequentially to the receptor, one per clock cycle. The outputs of the first $k$ shift registers from each register array become the input of the masking circuit, where padding is handled. The $k \times k$ outputs of the masking circuit become the output of the receptor.

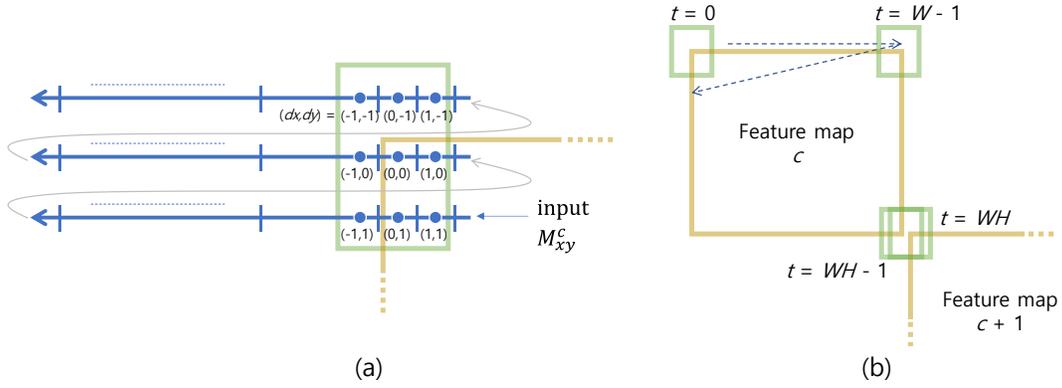

**Fig. 4. (a)** Receptor circuit consisting of $k$ shift register arrays, each with $W$ registers. The receptor converts one per clock cycle data flow to $k \times k$ per cycle. The outputs of the first $k$ shift registers of each register array become the output of the RU. **(b)** Right after the receptor scans the end of one image ($t = W \cdot H - 1$), the scanning of the next image begins ($t = W \cdot H$) without idle clock cycle.

**Table 1. Masking process for zero-padding ($k = 3$)**

| Clock cycle $t$ | Input to Receptor | Data on register arrays | | | Position values in the receptor ($x, y$) | | | Receptor output | | |
|---|---|---|---|---|---|---|---|---|---|---|
| $-(W \cdot \lfloor k/2 \rfloor + \lfloor k/2 \rfloor)$ | $d_{0 \cdot 0}$ (feature map 0) | | | $d_{0 \cdot 0}$ | | | | invalid | | |
| $-(W \cdot \lfloor k/2 \rfloor + \lfloor k/2 \rfloor) + 1$ | $d_{1 \cdot 0}$ | | $d_{0 \cdot 0}$ | $d_{1 \cdot 0}$ | | | | invalid | | |
| ... | | | | | | | | | | |
| 0 | $d_{1 \cdot 1}$ | | $d_{0 \cdot 0}$ | $d_{1 \cdot 0}$ | -1,-1 | 0,-1 | 1,-1 | 0 | 0 | 0 |
|   |   | $d_{W-1 \cdot 1}$ | $d_{0 \cdot 1}$ | $d_{1 \cdot 1}$ | -1,0 | **0,0** | 1,0 | 0 | $d_{0 \cdot 0}$ | $d_{1 \cdot 0}$ |
|   |   |   |   |   | -1,1 | 0,1 | 1,1 | 0 | $d_{0 \cdot 1}$ | $d_{1 \cdot 1}$ |
| 1 | $d_{2 \cdot 1}$ | $d_{0 \cdot 0}$ | $d_{1 \cdot 0}$ | $d_{2 \cdot 0}$ | 0,-1 | 1,-1 | 2,-1 | 0 | 0 | 0 |
|   |   | $d_{0 \cdot 1}$ | $d_{1 \cdot 1}$ | $d_{2 \cdot 1}$ | 0,0 | **1,0** | 2,0 | $d_{0 \cdot 0}$ | $d_{1 \cdot 0}$ | $d_{2 \cdot 0}$ |
|   |   |   |   |   | 0,1 | 1,1 | 2,1 | $d_{0 \cdot 1}$ | $d_{1 \cdot 1}$ | $d_{2 \cdot 1}$ |
| ... | | | | | | | | | | |
| $W - 1$ | $d_{0 \cdot 2}$ | | | $d_{0 \cdot 0}$ | $W$-2,-1 | $W$-1,-1 | $W$,-1 | 0 | 0 | 0 |
|   |   | $d_{W-2 \cdot 0}$ | $d_{W-1 \cdot 0}$ | $d_{0 \cdot 1}$ | $W$-2,0 | **$W$-1,0** | $W$,0 | $d_{W-2 \cdot 0}$ | $d_{W-1 \cdot 0}$ | 0 |
|   |   | $d_{W-2 \cdot 1}$ | $d_{W-1 \cdot 1}$ | $d_{0 \cdot 2}$ | $W$-2,1 | $W$-1,1 | $W$,1 | $d_{W-2 \cdot 1}$ | $d_{W-1 \cdot 1}$ | 0 |
| $W$ | $d_{1 \cdot 2}$ | | $d_{0 \cdot 0}$ | $d_{1 \cdot 0}$ | -1,0 | 0,0 | 1,0 | 0 | $d_{0 \cdot 0}$ | $d_{1 \cdot 0}$ |
|   |   | $d_{W-1 \cdot 0}$ | $d_{0 \cdot 1}$ | $d_{1 \cdot 1}$ | -1,1 | **0,1** | 1,1 | 0 | $d_{0 \cdot 1}$ | $d_{1 \cdot 1}$ |
|   |   | $d_{W-1 \cdot 1}$ | $d_{0 \cdot 2}$ | $d_{1 \cdot 2}$ | -1,2 | 0,2 | 1,2 | 0 | $d_{0 \cdot 2}$ | $d_{1 \cdot 2}$ |
| ... | | | | | | | | | | |
| $W \cdot H - 1$ | $d_{0 \cdot 1}$ (feature map 1) | $d_{W-2 \cdot H-2}$ | $d_{W-1 \cdot H-2}$ | $d_{0 \cdot H-1}$ | $W$-2,$H$-2 | $W$-1,$H$-2 | $W$,$H$-2 | $d_{W-2 \cdot H-2}$ | $d_{W-1 \cdot H-2}$ | 0 |
|   |   | $d_{W-2 \cdot H-1}$ | $d_{W-1 \cdot H-1}$ | $d_{0 \cdot 0}$ | $W$-2,$H$-1 | **$W$-1,$H$-1** | $W$,$H$-1 | $d_{W-2 \cdot H-1}$ | $d_{W-1 \cdot H-1}$ | 0 |
|   |   | $d_{W-2 \cdot 0}$ | $d_{W-1 \cdot 0}$ | $d_{0 \cdot 1}$ | $W$-2,$H$ | $W$-1,$H$ | $W$,$H$ | 0 | 0 | 0 |
| $W \cdot H$ | $d_{1 \cdot 1}$ | $d_{W-1 \cdot H-2}$ | $d_{0 \cdot H-1}$ | $d_{1 \cdot H-1}$ | -1,-1 | 0,-1 | 1,-1 | 0 | 0 | 0 |
|   |   | $d_{W-1 \cdot H-1}$ | $d_{0 \cdot 0}$ | $d_{1 \cdot 0}$ | -1,0 | **0,0** | 1,0 | 0 | $d_{0 \cdot 0}$ | $d_{1 \cdot 0}$ |
|   |   | $d_{W-1 \cdot 0}$ | $d_{0 \cdot 1}$ | $d_{1 \cdot 1}$ | -1,1 | 0,1 | 1,1 | 0 | $d_{0 \cdot 1}$ | $d_{1 \cdot 1}$ |
| ... | | | | | | | | | | |

The masking circuit manages the $x$ and $y$ position values of each of the $k \times k$ points as the clock cycle progresses. If we denote $t = 0$ as the time when $k \cdot W + \lfloor k/2 \rfloor$ clock cycles after the first data is fed to the input of the receptor, the position values of the center point are set as $x = t \% W$, $y = \lfloor t / W \rfloor$. The position values of the other points are determined by adding the relative position values $dx, dy$ to the position of the center point. Then, the output values of the data points with $x < 0$, $x \geq W$, $y < 0$, or $y \geq H$ are set to 0. These correspond to zero-padded inputs. Table 1 shows how data are processed in the masking circuit.

The outputs of the receptors become the output of the RU and constitute the input data of the HN described by Equation 3. Figure 4-b shows how feature maps are processed in sequence. After the last data of a feature map is output, the first data of the next feature map is output in the next clock cycle.

When computing the CNN layer with $k = 1$, the input and the output of the RU are the same, and the input data is bypassed to the output.

**Multiple HNs:** Instead of using a single HN, multiple HNs can be used to compute multiple feature maps in parallel. When $Q$ HNs compute $Q$ output feature maps simultaneously, the $f^{th}$ output feature map is computed by the $(f \% Q)^{th}$ HN as shown in Figure 1-(b). In this case, the MP stores $Q$ output feature maps simultaneously, and the output of the MP is replicated and provided to all HNs. $Q$ HNs can perform $Q \times P \times k \times k$ multiplications every clock cycle, and reduce the computation time for a CNN layer from $\lceil C/P \rceil \times F \times W \times H$ clock cycles to $\lceil C/P \rceil \times \lceil F/Q \rceil \times W \times H$. By selecting sufficiently large values of $Q$ and $P$, the design can be upscaled maintaining high efficiency.

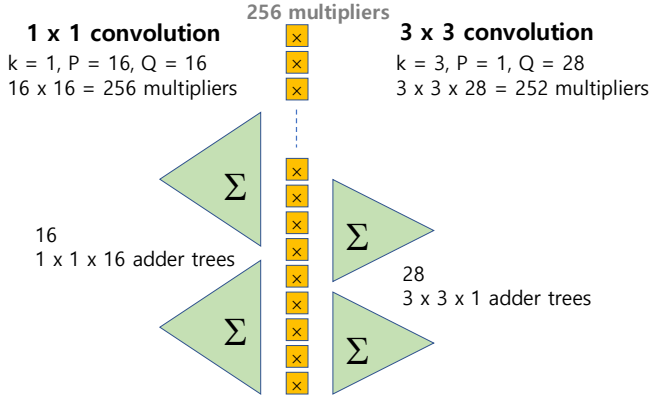

Fig. 5. Example of the shared multiplier/adder scheme. m = 256. When $k = 3$, $P = 1$ and $Q = 28$, and when $k = 1$, $P = 16$ and $Q = 16$. Switches in the circuit change the paths of the data before starting each CNN layer to reshape the circuit.

**Sharing multipliers for different configurations:** The convolution models use different filter sizes for their convolution layers. Typically, $k = 3$ and $k = 1$ are often used alternately. In this case, different numbers of multipliers and adders are required for the HN. Therefore, it is efficient to share a pool of multipliers and adders for the different configurations. For that purpose, a large number of multipliers and adders are deployed in the circuit. The configuration parameters of $k$, $P$, and $Q$ are designated differently based on the size of the CNN layer that is being computed, thereby flexibly utilizing the multipliers and adders. Using digital switches embedded in the circuit and controlled by the control unit (CU), the adders form $Q$ adder trees each sums $k \times k \times P$ inputs. For example, 252 of 256 adders ($m = 256$) can be used to form 28 adder trees each with 9 inputs when $k = 3$, $P = 1$ and $Q = 28$. Then, all 256 adders are used to form 16 adder trees each with 16 inputs when $k = 1$, $P = 16$, and $Q = 16$, as shown in Figure 5.

**Stored-Program Control Scheme:** The system control is relatively simple. Principally, using the counters that increment with each clock-cycle, the CU provides the read and write addresses of each memory to match with the pipeline latency. The starting address of all read ports of the MAU's memories is specified in the SOT. For the addresses of the read and write ports of the DU's Netsum memories, the data index of the input feature map is provided with a time difference. For each address, a new partial sum is written, after reading the previous partial sum. The starting address of all write ports of the MAU's memories is also specified in the SOT. Since the same weight data are applied to all elements of input feature maps, the read address of all weight memories in the SNU is increased by 1 every $W \cdot H$ clock cycles until the inference computation is over.

The proposed hardware architecture can compute convolution layers of various configurations so that multiple CNN models can be computed without changing the hardware design. For this, the control signals for the operation of the $l^{th}$ convolution layer are stored in the $l^{th}$ row of a table called the stage operation table (SOT). The control signals include system parameters such as the values of $W$, $H$, and $k$, and the read and write addresses of the memories in the MAU. As the reference time of each control point in the circuit differs owing to pipeline delay, each control signal from the SOT is also delayed by using pipeline registers to synchronize the control timing. Upon completion of all the steps in the SOT, that is, after all the CNN layers have been processed, the same procedure is repeated from the first row of the SOT for the next input image. The architecture runs by itself, and no processor is involved in the computation.

Such a scheme can be viewed as a stored-program system, where each row of the SOT corresponds to one very complex instruction set computer (CISC) instruction code which runs hundreds of thousands of clock cycles for a corresponding convolution layer. Changing the contents of the SOT allows it to run completely different CNN models.

IV. EVALUATION

A. Efficiency Metrics

Without a clear measure of efficiency, efficient computational systems cannot be designed. For this reason, we used two metrics in designing the system: the multiplier composition efficiency, $R_c$, and the multiplier utilization efficiency, $R_u$. They are defined by the following formulas:

$$R_c = \frac{resource\_used\_for\_multipliers}{all\_system\_resources} \quad (7)$$

$$R_u = \frac{actual\_multiplication\_speed}{peak\_multiplication\_speed} \quad (8)$$

$$= \frac{model\_speed \times model\_multiplication\_requirement}{peak\_multiplication\_speed} \quad (9)$$

$$Eff_{arch} = R_c \times R_u \quad (10)$$

For $R_c$ to be valid, the multipliers used in the system must be typical ones with no special restrictions imposed by architecture.

In the ideal system, both metrics are 1.0 and thus $Eff_{arch} = 1.0$. This means that all resources in the system are used only for implementing multipliers, and all such multipliers are fully utilized without an idle clock cycle. Multipliers cannot be removed to achieve a speed because multiplications are required by computational model. By definition, no system can have an efficiency higher than that of this ideal system. Figure 6-a depicts the efficiency metrics of the computational architecture. $Eff_{arch}$ is equal to the ratio of the green areas to the total resource area. Figure 6-b indicates an ideal efficiency.

Note that these efficiency metrics are purely architectural and are independent of chip technology and computational precision.

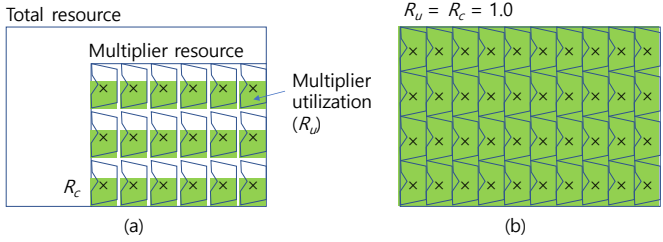

Fig. 6. (a) Implication of the efficiency metrics, $Eff_{arch} = R_c \times R_u$. (b) Ideal efficiency where both $R_u$ and $R_c$ are 1.0.

Table 2. SSD/MobileNet execution statistics of reference implementation

| L | Type | Input map size (W, H) | Output map size (W, H) | C | F | Multiplication requirement (A) | Number of clock cycles (B) | Peak multiplications (C = B x 256) |
|---|---|---|---|---|---|---|---|---|
| 1 | 3x3 | 300 | 150 | 3 | 32 | 19382400 | 135207 | 34612992 |
| 2 | DW3x3 | 150 | 150 | 32 | 1 | 6460800 | 45207 | 11572992 |
| 3 | 1x1 | 150 | 150 | 32 | 64 | 46080000 | 180207 | 46132992 |
| 4 | DW3x3 | 150 | 75 | 64 | 1 | 3220800 | 17007 | 4353792 |
| 5 | 1x1 | 75 | 75 | 64 | 128 | 46080000 | 180132 | 46113792 |
| 6 | DW3x3 | 75 | 75 | 128 | 1 | 6441600 | 28257 | 7233792 |
| 42 | 1x1 | 3 | 3 | 256 | 24 | 55296 | 348 | 89088 |
| 43 | 1x1 | 3 | 3 | 256 | 546 | 1257984 | 5100 | 1305600 |
| 44 | 1x1 | 2 | 2 | 256 | 24 | 24576 | 187 | 47872 |
| 45 | 1x1 | 2 | 2 | 256 | 546 | 559104 | 2299 | 588544 |
| 46 | 1x1 | 1 | 1 | 128 | 24 | 3072 | 74 | 18944 |
| 47 | 1x1 | 1 | 1 | 128 | 546 | 69888 | 338 | 86528 |
| | | | Total | | | 1.233x10$^9$ | 4958821 | 1.269x10$^9$ |
| | | | | | | Multiplier utilization rate ($R_u$) = A / C = | | 0.972 |

### B. Reference Implementation

We implement a system using the proposed design method and run the SSD/MobileNet object detection model [3] to evaluate its efficiency. The SSD/MobileNet CNN model, developed by Google, is the base model of the TensorFlow object detection environment and is one of the most popular CNN models. Execution results of SSD/MobileNet for various hardware are available. This model comprises 47 convolution layers and requires approximately 1.23 billion multiplications to inference a $300 \times 300$ image once. The model is composed of $3 \times 3$, depthwise (DW) $3 \times 3$, and $1 \times 1$ convolution layers.

Some computation statistics have been shown in Table 2.

### C. Multiplier Utilization Rate

The reference system computes the SSD/MobileNet model at a speed of 40.3 fps using 256 multipliers with a 200 MHz clock frequency. We use the same configuration of $P$ and $Q$ as exemplified in Figure 5. In addition, 32 memories are used for the MAU ($R = 32$).

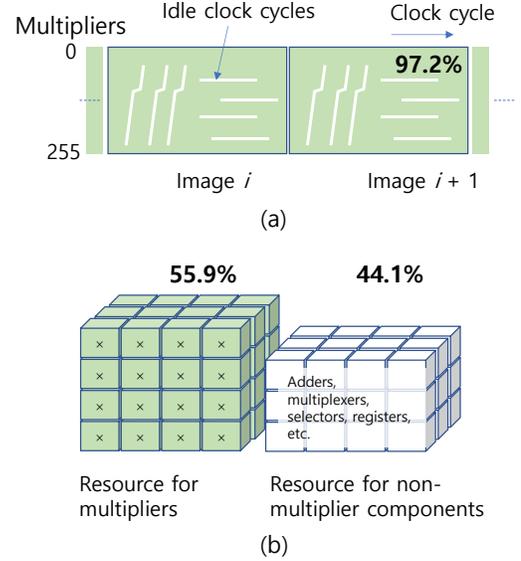

Fig. 7. (a) Conceptual diagram of the multiplier output footprint; in our reference system, 97.2% of all clock cycles of all the 256 multipliers produced effective results during the entire computation cycle. (b) Resource composition of the reference system. 55.9% of the total resources are used to implement 256 multipliers which are indispensable components.

A total of $4.96 \times 10^6$ clock cycles are taken to process each image. As the system can perform 256 multiplications in each clock cycle, the total number of multiplications performed by all the multipliers is $1.269 \times 10^9$. However, the effective number of multiplications required for the inference of the CNN model is $1.233 \times 10^9$; therefore, the utilization rate of the multipliers, $R_u$, is 97.2%, as shown in Table 2. This means that only 2.8% of all multiplier clock cycles are idle or ineffective over the entire computation cycle, as depicted in Figure 7-a.

The largest contribution in the 2.8% overhead is from the internal fragmentation (2.3%). This occurs while computing the $3 \times 3$ convolutions, when the output feature maps are assigned to the 28 HNs and some are left empty at the end. The second largest contribution is from the padding overhead (0.3%), which is caused owing to the zero padded input. The third is of the external fragmentation overhead, wherein $3 \times 3 \times 28 = 252$ multipliers are used out of 256 while computing the $3 \times 3$ convolution (0.14%).

### D. Multiplier Composition Rate

Xilinx's Artix-7 field-programmable gate array (FPGA) is used in this reference implementation. FPGA chips implement digital circuits using lookup tables (LUTs) instead of transistors. FPGA has an advantage that the contents of the tables can be reconfigured and reused. This Artix-7 is a low-end chip containing a small amount of LUT resources. Our reference implementation uses 43% of the chip's total LUT resources.

Here, we calculate the proportion of the system's total LUT resources used for the multipliers. This calculation is performed to estimate the proportion of the transistors of the multipliers to the total number of transistors in the system in case this design is ported to an application-specific integrated circuit (ASIC). Because the reference design uses a number of dedicated multipliers, the resources of the multipliers are replaced with the number of LUT resources required when each multiplier is implemented using LUTs, and the entire system is expanded to a circuit implemented only by LUTs. The calculation does not include the processors that do not participate in the CNN computation. The additional units for the CNN model such as softmax (for classification), prediction layer, bounding box computation circuit, NMS (for object detection), system framework, and control signal generation circuits are included.

The number of LUTs for each unit is provided in Table 3.

**Table 3. LUT resource utilization by system functions**

| Unit | System resource including multipliers (A) | Multiplier resource (B) |
|---|---|---|
| SNU | 50432 | 47104 |
| DU | 13979 | |
| SU | 2300 | |
| Softmax | 2435 | |
| SSD_box | 1787 | |
| NMS | 909 | |
| Framework and control | 12454 | |
| Total | 84296 | 47104 |
| Multiplier composition rate ($R_c$) = B / A = | | **0.559** |

A total of 55.9% of the LUTs in the system is utilized for the 256 multipliers in the SNU. This means that 55.9% of the total LUT resources that make up the reference system are used to implement the multipliers and 44.1% are used for components other than the multipliers, as shown in Figure 7-b. When implementing this design on an ASIC-based chip, as all components are expected to be converted to transistors with similar ratios, the multiplier resource ratio will also be similar to that of an FPGA.

Therefore, our reference implementation achieved $Eff_{arch} = R_u \times R_c = 0.533$.

*E. Efficiency comparison*

Table 4 compares the multiplier utilization rate, $R_u$, of the proposed design with other AI systems. Existing AI chips show lower $R_u$ results as compared to our design.

**Table 4. Multiplier utilization rate ($R_u$) comparison**

| AI hardware | Peak multiplications | Model speed (SSD/MobileNet) | Actual multiplications speed | $R_u$ |
|---|---|---|---|---|
| Jetson Javier [4][12] | 1.60E+13 | 665 | 8.18E+11 | 5.1% |
| Jetson TX2 [13][14] | 1.33E+12 | 11.5 | 1.41E+10 | 1.1% |
| Jetson Nano [15] | 4.72E+11 | 39 | 4.80E+10 | 10.2% |
| Coral Edge [16] | 1.97E+12 | 90.9 | 1.20E+11 | 6.1% |
| Intel NCS2 [17][18] | 5.00E+11 | 8.6 | 1.06E+10 | 2.1% |
| **Our Design** | **5.12E+10** | **40.3** | **4.97E+10** | **97.2%** |

This comparison is based on the assumption that all the components in the circuit use similar proportions of resources regardless of whether the design is implemented as an FPGA or an ASIC. This is supported by the data from [12][19][20][21], where non-multiplier components such as adders and multiplexers can be implemented with similar transistors per LUT, as in the migration from FPGA to ASIC.

It is difficult to directly compare the multiplier composition rates between AI chips, as there is not enough information about the multipliers implemented in those chips. However, estimations show that conventional AI chips have far more resources than the minimally required multiplier resources, implying very low $R_c$. For example, the Jetson Xavier chip is composed of 9 billion transistors and contains approximately twelve thousand 8-bit integer multipliers [4][12]. As the same type of multiplier can be implemented with 3,000 transistors [12][26], and the ideal system consisting only of multipliers described above can be implemented using $3.6 \times 10^7$ transistors. Therefore, in an extreme point of view, it can be seen that the Jetson Xavier chip uses 250 times more transistors than the minimum number of transistors needed.

It must be noted that most other AI hardware systems are designed for general purposes, and therefore the comparison would be valid only when they are used for dedicated purposes. In addition, data shown here estimates only the efficiency of the computer architecture in the chips, not their efficiencies. These chips provide good performance by using excellent semiconductor integration technologies.

V. COMMERCIAL APPLICATION

The reference design described in this paper is used for a commercial product called Deep Runner (Figure 8). It aims to apply AI to various industries by using image classification and object recognition algorithms to recognize image frames in video cameras.

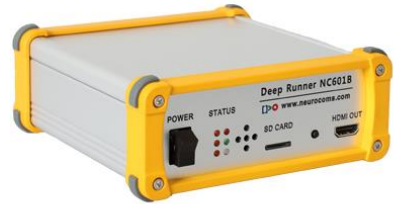

(a)

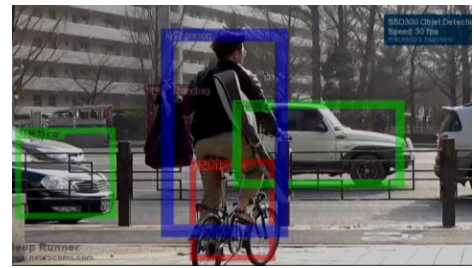

(b)

**Fig. 8. (a) Deep Runner NC601. (b) Object detection by Deep Runner**

Deep Runner supports Inception V1 (GoogleNet), MobileNet V2 1.4x classification models, YOLO, Tiny YOLO, and SSD/MobileNet object detection models. The main application area of Deep Runner NC601 is quality inspection in factories. It provides the functionality in recognizing high resolution product images, and general-purpose input/output port for pin-to-pin communication with programmable logic controllers controlling production lines. Deep Runner NC901 is targeted for security systems, and it can simultaneously recognize up to 16 CCTV cameras. In addition, it supports object counting, vehicle license plate recognition, and the web interface that automatically sends notifications when preset conditions are met.

For deep learning training of user-specific objects, Neurocoms's Deep Trainer software [27], and Google's TensorFlow [22] can be used.

## VI. DISCUSSION

From the definitions of $R_c$ and $R_u$, $Eff_{arch}$ cannot exceed 1.0. Therefore, our system's $Eff_{arch}$ which is greater than 0.5 indicates that no system can be more than twice as efficient as ours in terms of performance-to-resource ratio.

The high computational efficiency of this structure is achieved purely by improving the flow of data without premising of chip technology. Therefore, the same principle could be applied to completely different physical systems, such as quantum computing or ternary systems.

In this work, we have described the architecture that performs inference only. However, this scheme can also be applied to a system that supports deep learning training. Moreover, backpropagation, the key algorithm for training, can be implemented based on the neuron machine architecture as shown in [23].

## VII. CONCLUSION

In this work, we have proposed a novel computational architecture for CNN that achieves high computational efficiency regardless of chip technology. Our reference system uses most of the peak multiplication capability in actual CNN computations, and the proportion of additional components other than multipliers in total system resources is no greater than multipliers. If the proposed design is implemented with the latest chip technology, it could achieve the same speeds using only a small fraction of transistors as compared to existing AI chips. We showed that the efficiency of this design is close to that of an ideal system, whose efficiency cannot be improved further.